\documentclass[letterpaper, 10 pt, conference]{ieeeconf}
\IEEEoverridecommandlockouts
\usepackage{cite}
\usepackage{amsmath,amssymb,amsfonts}
\usepackage{algorithmic}
\usepackage{graphicx}
\usepackage{textcomp}
\usepackage{colortbl}
\usepackage{float}
\usepackage{booktabs}
\usepackage[table]{xcolor}
\definecolor{maroon}{rgb}{0.6, 0.87, 0.94}
\definecolor{awesome}{rgb}{0.98, 0.66, 0.68}
\usepackage{multirow}
\usepackage{etoolbox}
\usepackage{setspace}
\usepackage{etoolbox}
\usepackage{multirow}
\usepackage{algorithmic}
\usepackage{algorithm}
\usepackage{threeparttable}

\usepackage{hyperref}
\makeatletter

\title{\LARGE \bf
VLA Model-Expert Collaboration for Bi-directional \\ Manipulation Learning}

\begin{document}

\author{Tian-Yu Xiang$^\dagger$, Ao-Qun Jin$^\dagger$, Xiao-Hu Zhou$^*$, Mei-Jiang Gui, Xiao-Liang Xie, Shi-Qi Liu, \\Shuang-Yi Wang, Sheng-Bin Duang, Si-Cheng Wang, Zheng Lei, Zeng-Guang Hou$^*$
\thanks{This work was supported in part by the National Key Research and Development Program of China under 2023YFC2415100, in part by the National Natural Science Foundation of China under Grant 62222316, Grant 62373351, Grant 82327801, Grant 62073325, Grant 62303463, in part by the Chinese Academy of Sciences Project for Young Scientists in Basic Research under Grant No.YSBR-104 and in part by China Postdoctoral Science Foundation under Grant 2024M763535.}
\thanks{The authors are with State Key Laboratory of Multimodal Artificial Intelligence Systems, Institute of Automation, Chinese Academy of Sciences, and also with School of Artificial Intelligence, University of Chinese Academy of Sciences, Beijing 100049, China. E-mail: $\{$xiangtianyu2021, xiaohu.zhou, zengguang.hou$\}$@ia.ac.cn}
\thanks{$\dagger$ Equally contribution: Tian-Yu Xiang, Ao-Qun Jin, $*$ Corresponding author: Xiao-Hu Zhou, Zeng-Guang Hou.}
}

\maketitle
\thispagestyle{empty}
\pagestyle{empty}

\begin{abstract}

The emergence of vision-language-action (VLA) models has given rise to foundation models for robot manipulation. Although these models have achieved significant improvements, their generalization in multi-task manipulation remains limited. This study proposes a VLA model-expert collaboration framework that leverages a limited number of expert actions to enhance VLA model performance. This approach reduces expert workload relative to manual operation while simultaneously improving the reliability and generalization of VLA models. Furthermore, manipulation data collected during collaboration can further refine the VLA model, while human participants concurrently enhance their skills. This bi-directional learning loop boosts the overall performance of the collaboration system. Experimental results across various VLA models demonstrate the effectiveness of the proposed system in collaborative manipulation and learning, as evidenced by improved success rates across tasks. Additionally, validation using a brain-computer interface (BCI) indicates that the collaboration system enhances the efficiency of low-speed action systems by involving VLA model during manipulation. These promising results pave the way for advancing human-robot interaction in the era of foundation models for robotics. (Project website: \href{https://aoqunjin.github.io/Expert-VLA/}{https://aoqunjin.github.io/Expert-VLA/})

\end{abstract}

\begin{keywords}
Human-Robot Collaboration; Human Factors and Human-in-the-Loop; Learning from Demonstration
\end{keywords}

\section{Introduction}

Motivated by the successful application of large-scale data to enhance generalization and robustness in computer vision~\cite{kirillov2023segment} and natural language processing~\cite{achiam2023gpt}, recent efforts in robot learning have focused on leveraging extensive manipulation data to develop robotic foundation models~\cite{vuong2023open, kim2024openvla, mees2024octo, bousmalis2023robocat}. These studies design algorithms trained on diverse tasks, environments, and robotic embodiments, aiming to develop generalized policies across settings and platforms. Beyond dataset scale, robotic foundation models incorporate principles from vision and language models, using language instructions—processed via pre-trained models such as LLaMA 2~\cite{touvron2023llama}—to guide manipulation, while vision models like SigLIP ViT~\cite{zhai2023sigmoid} condition action sequences on visual inputs. These architectures, termed \textbf{V}ision-\textbf{L}anguage-\textbf{A}ction (VLA) models~\cite{zitkovich2023rt}, integrate prior task knowledge from vision and language, surpassing traditional robotic learning approaches~\cite{ma2024survey}.

\begin{figure}[tb]
   \centering
     \includegraphics[width=3.5in]{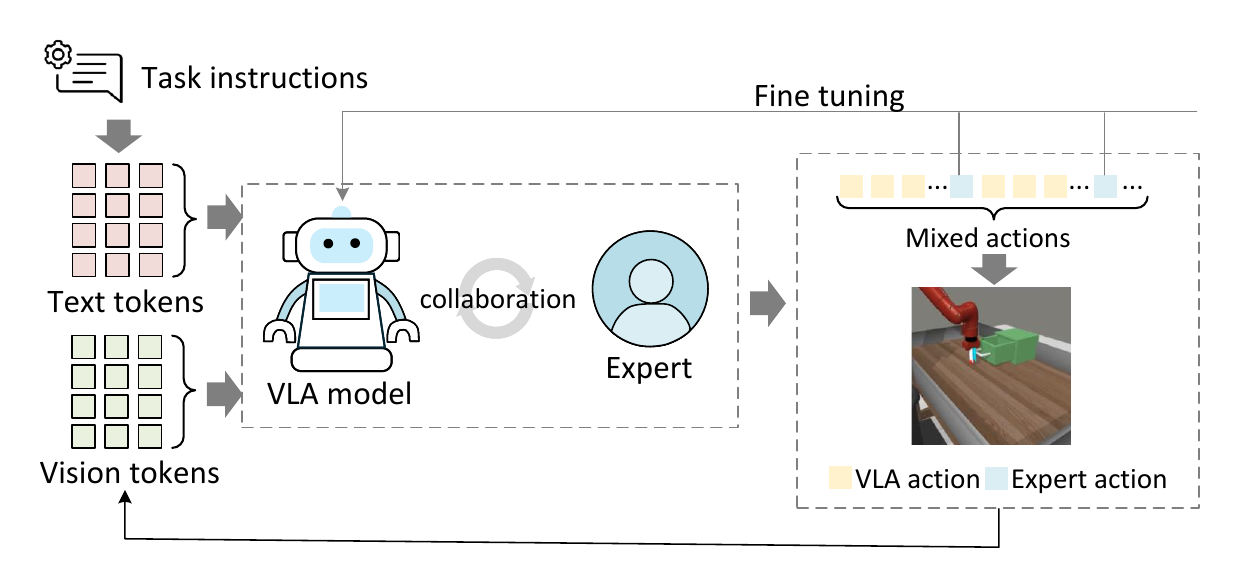}
       \caption{The proposed VLA model-expert collaboration system integrates a VLA model and expert interactions to enhance manipulation. The VLA model generates actions by processing task instructions as text tokens and environmental inputs as vision tokens. Meanwhile, the expert makes decisions at a lower frequency, assisting the VLA model. Expert-executed actions are collected to fine-tune the VLA model, improving system performance.}
    \centering
    \label{fig:intro}
\end{figure}

Although VLA models have advanced autonomous manipulation abilities over traditional robotic learning algorithms, challenges persist in developing universe policy across the environment due to the scarcity of high-quality demonstrations and the diversity of manipulation tasks. Compared to computer vision and natural language processing, the scale of robotic manipulation datasets remains relatively limited. For example, the Open X-Embodiment dataset~\cite{vuong2023open}, the largest open-source robotic manipulation dataset, contains approximately 2.5 million demonstrations—even significantly fewer than pre-large-model era datasets in other domains, such as ImageNet with over 10 million images and GPT-2, trained on 8 million web pages~\cite{deng2009imagenet, radford2019language}. Furthermore, manipulation tasks exhibit greater heterogeneity and abstraction than vision and language processing. Unlike perception tasks, manipulation skills are inherently difficult to transfer, even for biological intelligence such as humans. In addition, a single manipulation task can be executed through multiple strategies, expanding the state space in policy learning.

To effectively deploy VLA models in target environments, their manipulation capabilities need to be enhanced for downstream tasks~\cite{vuong2023open, dalal2023imitating, black2024pi_0, bousmalis2023robocat, guo2025improving, carta2023grounding}. One approach is fine-tuning VLA models with task-specific manipulation data, a widely adopted strategy in large-model. Pre-trained VLA models across multiple embodiments and environments exhibit positive transfer in downstream application, as evidenced by improved performance on target tasks~\cite{vuong2023open, dalal2023imitating}. Another approach involves integrating expert decisions with the policy model to semi-autonomous systems. By delegating limited actions to experts and assigning the majority of routine operations to the policy model, this collaboration reduces the expert workload while mitigating the limitations of the VLA models in complex cases. Although expert-in-the-loop frameworks are common in semi-autonomous robotics~\cite{wang2021predicting, ma2024human}, their integration with VLA models is still an open problem.

To overcome these limitations, this study integrates expert-in-the-loop and fine-tuning techniques to enable collaborative manipulation and learning between experts and VLA models. The VLA model is first fine-tuned with a small amount of task-specific data, followed by collaboration with experts to accomplish target tasks. During the collaboration, human experts become more familiar with the system and more skillful in manipulation. Manipulation data collected during VLA model-expert collaboration is stored in a buffer for subsequent fine-tuning, enabling continuous performance improvement (See Fig.~\ref{fig:intro}). The contributions of this study can be summarized as follows:

\begin{itemize}
    \item Semi-autonomous manipulation is achieved via collaboration between the VLA model and experts. To the best of our knowledge, this work is a pioneer study in investigating VLA model-expert collaboration.
    \item The collaborated process enables bi-directional learning: VLA models can be further fine-tuned using manipulation data, while experts adapt to the VLA model.
    \item Experimental results in the MetaWorld environment confirm the effectiveness of collaboration. With an action ratio of the VLA model to the expert set to $4:1$, the success rate of the VLA model in different tasks improves by $6.2\%/13.5\%$ for MT10/MT50 benchmark, and the number of action steps for human experts decreases by $82.24\%$.
\end{itemize}

\section{Related Works}

\subsection{VLA Models for Robot Learning}

Building on the success of vision and language foundation models, VLA models have emerged as a promising approach for developing generalist robot policies~\cite{li2024towards, ma2024survey}. These models leverage visual and language representations to provide high-level task instructions and contextual cues for low-level actions. 

Based on their input-output structures~\cite{li2024towards}, VLA models can be categorized into four types: One-Step input with Discrete-Action output (OSDA)~\cite{zitkovich2023rt, zhen3d, kim2024openvla}, Historical-Step input with Discrete-Action output (HSDA)~\cite{wuunleashing, brohan2022rt, cheang2024gr}, One-Step input with Continuous-Action output (OSCA)~\cite{zhaolearning, jang2022bc, radosavovic2023real, black2024pi_0}, and Historical-Step input with Continuous-Action output (HSCA)~\cite{mees2024octo, reed2022generalist, bousmalis2023robocat}. 

The key distinction between one-step and historical-step models is whether actions are predicted solely from the current observation or incorporate historical context. While robot action spaces are inherently continuous, the VLA model's action space can be either continuous or discretized, depending on the design of the action head. A discrete action head leverages the structure of the language decoder to discretize the continuous action space, assigning specific values to each token in the output layer, transforming action prediction into a classification problem~\cite{zitkovich2023rt, kim2024openvla, brohan2022rt}. In contrast, continuous actions can be generated using methods such as a diffusion-based head~\cite{cji2024diffusion, mees2024octo} or a flow-matching-based head~\cite{black2024pi_0}. This study evaluates representative VLA models from different categories within the proposed VLA model-expert collaboration system~\cite{mees2024octo, black2024pi_0, kim2024openvla}.

\subsection{Fine-tuning Techniques for Robot Manipulation Models}

Fine-tuning plays a crucial role in adapting pre-trained robot models to downstream applications~\cite{kim2024openvla, mees2024octo, vuong2023open, dalal2023imitating, black2024pi_0, bousmalis2023robocat}. The most straightforward approach involves fine-tuning the models with a limited number of target manipulation trials, which has shown effectiveness in several models~\cite{kim2024openvla, mees2024octo, vuong2023open, black2024pi_0}. Due to the challenges and time-consuming nature of data collection, self-improvement techniques have emerged. These techniques allow models to fine-tune using synthetic data generated by the model itself~\cite{dalal2023imitating, bousmalis2023robocat}. However, erroneous manipulations within the generated data can degrade model performance. Another fine-tuning approach involves reinforcement learning (RL), where high-level planning or low-level control policies are optimized using designed rewards~\cite{guo2025improving, carta2023grounding, szot2023large}. While effective, RL requires extensive agent-environment interactions, making it a challenge to deploy. This study proposes an alternative fine-tuning paradigm where expert interactions with the model serve as an optimal policy. By leveraging historical manipulation data during the collaboration with experts, the VLA model can refine its capabilities, improving performance through interaction. 

\begin{figure*}[tb]
   \centering
     \centering
     \includegraphics[width=7in]{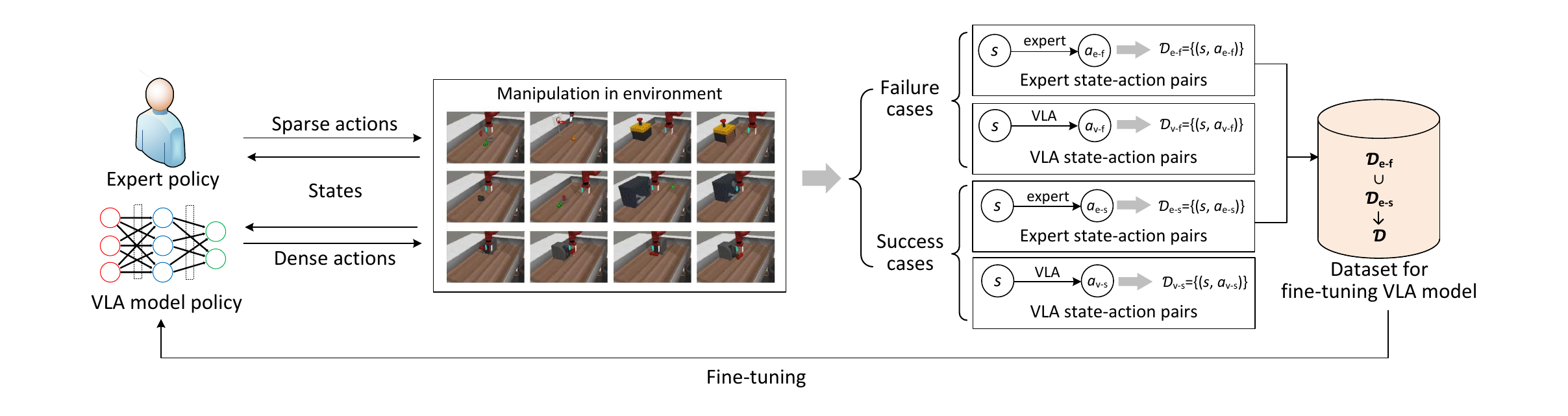}
       \caption{Collaboration pipeline between VLA model and expert for manipulation and learning.}
    \label{fig:pca}
\end{figure*}

\section{Method}

\subsection{General Structure of VLA models}

The VLA models take vision input and language instructions as conditioned states to predict actions, which can be represented as $\mathcal{V} \times \mathcal{L} \rightarrow \mathcal{A}$. Here, $\mathcal{V}$ represents the visual input space, $\mathcal{L}$ represents the language instruction set, and $\mathcal{A}$ represents the action space of the robot. The VLA model establishes a mapping between vision-language input and the action space.

As a multimodal model, the vision and language inputs are first encoded through separate encoders, each designed for its respective modality. The deep representations extracted by these encoders are then fused, either through networks like FiLM~\cite{perez2018film} or by directly concatenating. Since this study does not focus on the design of the VLA model, here a general form for the input-output relationship of VLA models is provided:

\begin{equation} 
a_i = \pi_\text{VLA}(l_i, v_i) 
\end{equation}
where $a_i$ is the action or action sequence generated by the VLA model, $l_i \in \mathcal{L}$ is the language instruction, $v_i \in \mathcal{V}$ is the visual input, and $\pi_\text{VLA}$ is the policy of the VLA model. The policy $\pi_\text{VLA}$ predicts the action at the $i$-th step based on the vision observation $v_i$ and language instruction $l_i$. Note that depending on the model's structure, the vision input may either include historical information or only the current observation.

Given the continuous nature of the robot action space, when the VLA model employs a discrete action head, the predicted discrete action must be mapped back to the original continuous space:

\begin{equation}
    a_i = \frac{\hat{a}_i}{\text{vocab\_size} - 1} (a_{\max} - a_{\min}) + a_{\min}
\end{equation}
where $\hat{a}_i$ denotes the discrete action value corresponding to the predicted token, $\text{vocab\_size}$ is the vocabulary size, and $a_{\max}$ and $a_{\min}$ represent the upper and lower bounds of the action, respectively. This transformation maps discrete actions back to the continuous action space.

\begin{algorithm}[tb]
    \caption{: Collaborated Learning}
    \begin{algorithmic}[1]
    \REQUIRE VLA model's policy: $\pi_{\text{VLA}}$, expert's policy $\pi_{e}$, fine-tuning steps: $S$, and collaboration epoch: $T$
        \STATE Load pre-trained VLA model, initialize a data buffer $\mathcal{D}_{\text{buffer}}$ and a dataset $\mathcal{D}$ for fine-tuning VLA.
        \FOR{$x = 1$ to $T$}
            \WHILE{Buffer $\mathcal{D}_{\text{buffer}}$ is not full}
                \STATE Carry out tasks based the collaborated manipulation pipeline with $\pi_{\text{VLA}}$ and $\pi_{e}$.
                \STATE Collect actions generated based on $\pi_{e}$ and corresponding vision/language states in to $\mathcal{D}_{\text{buffer}}$.
            \ENDWHILE
            \STATE Save $\mathcal{D}_{\text{buffer}}$ into dataset $\mathcal{D}$ and clear $\mathcal{D}_{\text{buffer}}$.
            \FOR{$s = 1$ to $S$}
                \STATE Sample mini-batch from $D$.
                \STATE Update $\pi_{\text{VLA}}$ with supervised learning paradigm.
            \ENDFOR
        \ENDFOR
    \end{algorithmic}
    \label{alg_2}
\end{algorithm}

\subsection{Expert Policy}

Two expert policies are considered in this study.

\subsubsection{Rule-based Policy}

The rule-based policy is implemented within the MetaWorld simulation environment~\cite{yu2020meta}. The policy is defined by the contributors, with the robot being aware of both the task goal and the targets' position. The robot directly controls the arm to reach the desired position. Given that the inverse dynamics of the robot are known, this policy provides a near-optimal solution for accomplishing the tasks.

\subsubsection{Human Users Policy}

The human users' policy involves participants controlling the robot arm during the task. Although participants are encouraged to manipulate the arm to reach the target, their varying proficiency, coupled with potential mismatches between the 2D and 3D environments, may affect performance. As a result, while the policy can achieve the target, it may not be optimal.

\subsection{Expert-VLA Collaboration}

The collaboration between VLA models and experts consists of two key processes: manipulation and learning. In the manipulation step, the expert policy assists VLA models in accomplishing various manipulation tasks. The learning step utilizes data collected during manipulation to fine-tune the VLA model, further enhancing its performance (See Fig.~\ref{fig:pca}).

\subsubsection{Collaborated Manipulation}

Given that VLA models can handle most manipulation tasks except under extreme conditions, this study incorporates a limited number of expert policy actions as a complement to the VLA model. The collaborated manipulation process follows a straightforward design: for a given task, the VLA model autonomously executes actions for $N$ steps, followed by one step from the expert policy, repeating this cycle until the task is either completed or reaches the failure threshold.

Since the majority of steps in a given manipulation task can be successfully completed by the VLA model, the proportion of actions generated by the VLA model can be higher than those from the expert policy (e.g., four times more). This collaborative approach enhances the performance compared to a VLA model-only policy while reducing the burden on user-driven manipulation.

\subsubsection{Collaborated Learning}

Through the interaction between the VLA model and the expert policy, new manipulation data can be collected and used for further fine-tuning of the VLA model. Despite failure cases occur during the collaborated manipulation stage, due to the low proportion of expert-policy actions failing to accomplish tasks. The expert policy in these cases can still be considered a near-optimal action at the corresponding steps. Consequently, this data is also used for fine-tuning the VLA model. The detailed implementation of the collaborative learning process is provided in Algorithm~\ref{alg_2}. It should be noted that the collaborative learning process is bidirectional. If the expert policy is provided by human participants, interaction with the system can also enhance the participant’s proficiency in manipulation.

\section{Experiments and Results}

The experiments are designed to validate the following hypotheses:
\begin{itemize}
    \item VLA models and experts benefit from the collaboration by enhancing the performance of VLA models and reducing the experts' workload. 
    \item VLA models and experts can adapt to manipulation tasks through collaboration, enabling bi-directional learning.
\end{itemize}

\begin{table}[tb] \small
\renewcommand{\arraystretch}{1.2} 
\centering
\caption{Structure of VLA models in this study}
\begin{tabular}{p{1.5cm}<{\centering} p{0.8cm}<{\centering} p{2cm}<{\centering} p{1.2cm}<{\centering}p{1cm}<{\centering}} 
\toprule
\textbf{Models} & \textbf{Type} & \textbf{V-model} & \textbf{L-model} & \textbf{Fusion} \\
\midrule
$\pi_0$~\cite{black2024pi_0} & OSCA & \multicolumn{3}{c}{PaliGemma (Vision-language model)~\cite{beyer2024paligemma}}  \\
OpenVLA~\cite{kim2024openvla} & OSDA & DINOv2~\cite{oquab2024dinov2} & SigLiP~\cite{zhai2023sigmoid} & Concat \\
Octo~\cite{mees2024octo} & HSCA & Light-CNN & T5~\cite{raffel2020exploring} & Concat \\

\bottomrule
\end{tabular}
\label{tbl:model}
\end{table}

\subsection{Implementation Details}

\subsubsection{VLA models}

In this study, three representative state-of-the-art models are selected to validate the proposed framework~\cite{black2024pi_0, kim2024openvla, mees2024octo}. These models differ in input (one-step/historical steps) and output (continuous/discrete action), representing different types and routes of VLA models (Table~\ref{tbl:model}).

\subsubsection{Experimental Environment}

The proposed framework is validated in the MetaWorld environment using the ML10 and ML50 benchmarks~\cite{yu2020meta}, which evaluate multi-task learning algorithms with 10 (ML10) or 50 (ML50) tasks. Since VLA models are designed as foundation models for manipulation tasks, they must handle a variety of tasks, not just a single one. Therefore, this study is evaluated under the multi-task paradigm. For each task, 50 trajectories are collected using a rule-based policy for fine-tuning the VLA models, with each trajectory limited to a maximum of 500 steps. Data are captured using a fixed camera setup with an elevation angle of -25° and an azimuth of 145°.

\subsubsection{Training Details}

A series of data augmentation techniques are applied to the vision inputs during fine-tuning to enhance model generalization. These augmentations include random resized cropping (90\% of the original size), random brightness adjustment (\(\pm20\%\)), random contrast adjustment (within \([0.8, 1.2]\)), random saturation adjustment (within \([0.8, 1.2]\)), and random hue adjustment (\(\pm0.05\)).

The action outputs of VLA models are normalized using min-max normalization for the discrete action model and z-score normalization for the continuous action model, following prior work~\cite{cji2024diffusion, mees2024octo}. The models are implemented based on the official code provided by the authors~\cite{perez2018film, mees2024octo, kim2024openvla}. Each model is fine-tuned with 800K sampled data from rule-based trajectories, optimized using the optimizer specified in the original paper or code.

\subsubsection{Evaluation}

During the evaluation stage, each task in MT10 or MT50 is tested 50 times with randomly initialized states. To ensure a fair comparison, all VLA models across different settings are evaluated using the same set of random seeds for the task initialization.

\begin{table}[!tb] \small
\renewcommand{\arraystretch}{1.2} 
\centering
\begin{threeparttable}
\caption{Comparison between Octo collaborated with rule-based expert policy and human expert policy in MT10 ($N=4$).}
\begin{tabular}{p{2cm} p{0.8cm}<{\centering} p{0.8cm}<{\centering} p{0.8cm}<{\centering} p{0.8cm}<{\centering} p{0.8cm}} 
\toprule

\multirow{2}{*}{\textbf{Tasks}} & \multicolumn{3}{c}{\textbf{Successful rate}} & \multicolumn{2}{c}{\textbf{Steps}} \\
\cmidrule{2-6}
& V & V-R & V-H  & H & V-H\\
\midrule
window open  & 1.00 & 1.00 & 0.98 & 85.96 & 18.65\\
reach & 0.34 & 0.32 & 0.86 & 68.98 & 12.07\\
peg insert & 0.30 & 0.18 & 0.52 & 157.86 & 28.42\\
drawer close & 1.00 & 1.00 & 1.00 & 42.94 & 16.42\\
drawer open & 0.92 & 1.00 & 0.96 & 68.54 & 21.33\\
push & 0.56 & 0.20 & 0.80 & 107.58 & 13.13\\
button press & 0.30 & 1.00 & 0.92 & 120.10 & 13.70\\
window close & 1.00 & 1.00 & 0.96 & 88.40 & 20.81\\
pick place & 0.54 & 0.46 & 0.58 & 112.87 & 14.21\\
door open & 0.96 & 1.00 & 1.00 & 148.00 & 19.08\\
\midrule
Average & 0.69 & 0.72 & 0.86 & 100.12 & 17.78 \\
\bottomrule
\label{tbl:Results2}
\end{tabular}

\begin{tablenotes}
    \item[1] `V': VLA model (Octo); `V-R': Collaboration between VLA model and rule-based expert policy; `V-H': Collaboration between VLA model and human expert policy; `H': Human expert policy.
\end{tablenotes}
\end{threeparttable}

\end{table}

\begin{table*}[!ht] \small
\renewcommand{\arraystretch}{1.2} 
\centering
\begin{threeparttable}
\caption{Success rate of collaboration between VLA models and rule-based expert policy under different ratios (VLA/expert) in MT10 and MT50 benchmarks.}
\begin{tabular}{p{1.5cm}<{\centering} p{2cm}<{\centering} p{1.1cm}<{\centering} p{1.1cm}<{\centering} p{1.3cm}<{\centering} p{1.1cm}<{\centering} p{1.1cm}<{\centering} p{1.1cm}<{\centering} p{1.1cm}<{\centering} p{1.1cm}<{\centering}} 
\toprule
\multirow{2}{*}{\textbf{Benchmark}} & \multirow{2}{*}{\textbf{Model}}  & \multirow{2}{*}{\textbf{Baseline}} & \multicolumn{6}{c}{\textbf{Expert-VLA Collaborated Manipulation}} \\
\cmidrule{4-9}
& &  & \textbf{$N=32$} & \textbf{$N=16$} & \textbf{$N=8$} & \textbf{$N=4$} & \textbf{$N=2$} & \textbf{$N=1$} \\
\midrule
\multirow{4}{*}{MT10} & $\pi_0$ \cite{black2024pi_0} & 0.754 & 0.746 & 0.758 & 0.788 & 0.846 & 0.892 & 0.948 \\
& OpenVLA~\cite{kim2024openvla} & 0.854 & 0.862 & 0.892 & 0.904 & 0.924 & 0.954 & 0.988 \\
& Octo~\cite{mees2024octo} & 0.692 & 0.680 & 0.676 & 0.698 & 0.716 & 0.822 & 0.824 \\
\cmidrule{2-9}
& \multicolumn{2}{c}{Improvement}  & -0.004 & +0.008 & +0.030 & +0.062 & +0.122 & +0.153 \\ 
\midrule
\multirow{4}{*}{MT50} & $\pi_0$ \cite{black2024pi_0} & 0.566 & 0.568 & 0.601 & 0.651 & 0.728 & 0.808 & 0.918 \\ 
& OpenVLA~\cite{kim2024openvla} & 0.844 & 0.870 & 0.890 & 0.916 & 0.938 & 0.946 & 0.965 \\
& Octo~\cite{mees2024octo} & 0.446 & 0.471 & 0.495 & 0.539 & 0.594 & 0.658 & 0.756 \\
\cmidrule{2-9}
& \multicolumn{2}{c}{Improvement} & +0.018 & +0.043 & +0.083 & +0.135 & +0.185 & +0.261 \\ 
\bottomrule
\end{tabular}
\label{tbl:Results}
\begin{tablenotes}
    \item[1] Baseline models are initialized with pre-trained weights and fine-tuning following the implementation details.
    \item[2] Improvement: average improvement from the baseline (Only VLA model manipulation).
\end{tablenotes}
\end{threeparttable}

\end{table*}

\begin{figure}[tb]
   \centering
     \includegraphics[width=3.5in]{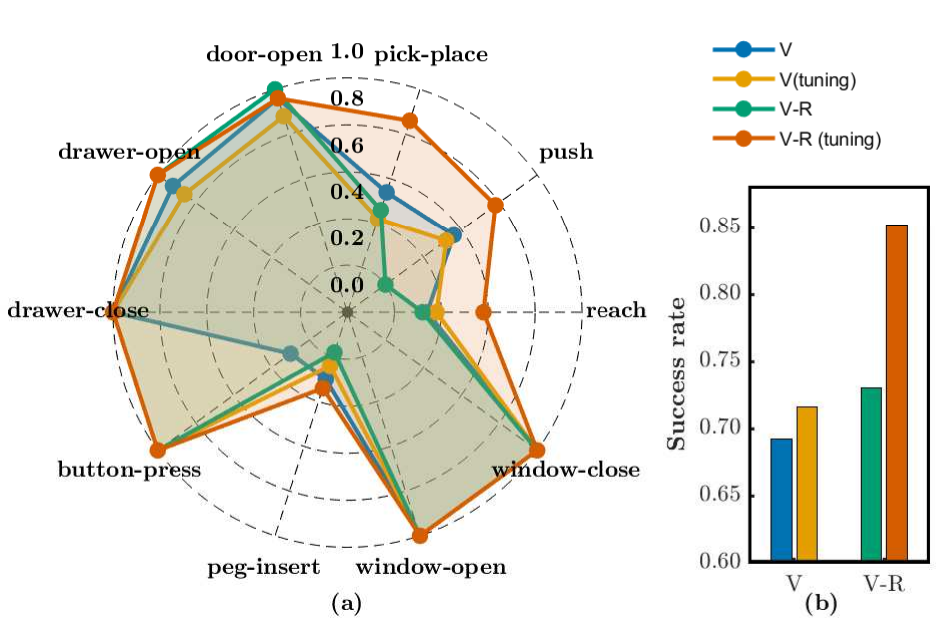}
       \caption{Comparison of the baseline VLA model (Octo) and the VLA model after collaborative learning (tuning). The success rates of the fine-tuned VLA model—with and without the rule-based expert policy (V vs. V-R)—are presented at the task level (a) and at the average level (b) in the MT10 benchmark.}
    \centering
    \label{fig:VLA_learning}
\end{figure}

\begin{figure}[tb]
   \centering
     \includegraphics[width=3.5in]{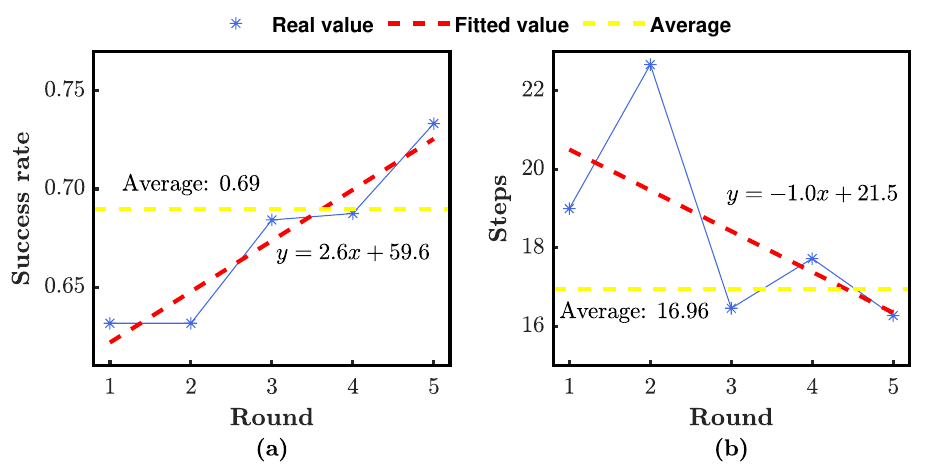}
       \caption{Visualization of success rate and action steps executed by human expert over round in hard tasks (successful rate lower than average) under MT10.}
    \centering
    \label{fig:human_learning}
\end{figure}

\subsection{Collaborated Manipulation}

\subsubsection{Successful Rate Improves for the VLA Models}

For VLA models~\cite{black2024pi_0, kim2024openvla, mees2024octo}, incorporating a small proportion of expert policy—either rule-based or human user policy—consistently improves success rates across different models in both MT10 and MT50 benchmarks. As shown in Table~\ref{tbl:Results}, increasing the ratio of expert actions based on rules enhances performance. This aligns with intuition, as expert policies generally outperform VLA policies; thus, a higher proportion of expert guidance leads to better outcomes.

It is noteworthy that although the success rate on the MT50 benchmark—which includes more manipulation tasks—is lower, the relative improvement is more pronounced. As the ratio of expert actions increases, the success rates across the two benchmarks converge (Baseline: 0.767 vs. 0.619; $N=1$: 0.920 vs. 0.880). This observation indicates that, even when the VLA model may fail to complete a task under larger manipulation sets, its actions are not entirely erroneous.

Furthermore, human user policies have been integrated with Octo to evaluate human-VLA collaboration Table~\ref{tbl:Results2}. Five participants interacted with Octo in real-time on the MT10 benchmark (5 participants $\times$ 10 times/task, aligning with rule-based expert experiments) under the setting of $N=4$ (four VLA model actions followed by one human expert action). The results show a significant improvement in task completion rate compared to the baseline model. Notably, human collaboration yielded better performance gains than rule-based policies, likely because the VLA model is fine-tuned using rule-based data. Human inputs, being more flexible and diverse, complement the original VLA model policy.

\subsubsection{Manipulation Steps Decrease by Human Users}

The benefits of collaboration for human users are evident in the reduction of workload, as reflected in the number of action steps executed by participants. This claim is validated by directly comparing action steps in a pure human-policy setting with those in human-VLA collaboration. Five participants performed tasks in the MT10 benchmark ($5$ participants $\times$ $10$ trials per task, following prior settings). The results are summarized in Table~\ref{tbl:Results2}. 

With a VLA-to-human action ratio of $4:1$ ($N=4$), the VLA model is expected to take about $80\%$ of the actions, leading to an equivalent reduction in human effort. The observed decrease ($82.24\%$) closely aligns with this expectation and slightly surpasses it. These findings suggest the collaboration not only reduces the number of human-executed actions but also enables the VLA model to sometimes take more optimal actions than human users.

\subsection{Collaborated Learning}

\subsubsection{VLA Models Improving through Historical Manipulation Data}

To test whether historical manipulation data can be used to fine-tune the VLA model, the Octo model is re-tuned using collaborated manipulation data with rule-based expert ($N=4$). As shown in Fig.~\ref{fig:VLA_learning}, the success rate of the VLA model improves for most manipulation tasks after re-tuning with collaboration data. The average success rate across different tasks increases by $0.038$ (from $0.692$ to $0.730$). This improvement indicates that the VLA model benefits from learning during collaboration, likely due to the expert policy guiding the model to complete tasks it could not previously accomplish. This manipulation data provides the VLA model with valuable examples of corner cases it would not otherwise encounter, thus enhancing its overall performance.

Another comparison evaluates the performance of collaborative manipulation before and after collaborative learning. An improvement in the average success rate is observed ($0.136$, from $0.716$ to $0.852$). Notably, after collaborative learning, the improvement due to collaborative manipulation is much higher ($0.122$ vs. $0.024$). This may be because, in cases where the VLA model fails when operating independently, re-tuning brings it closer to success. It may also imply that the enhancement in VLA performance is more pronounced than what is reflected by the success rate.

\begin{figure}[tb]
   \centering
     \includegraphics[width=3.5in]{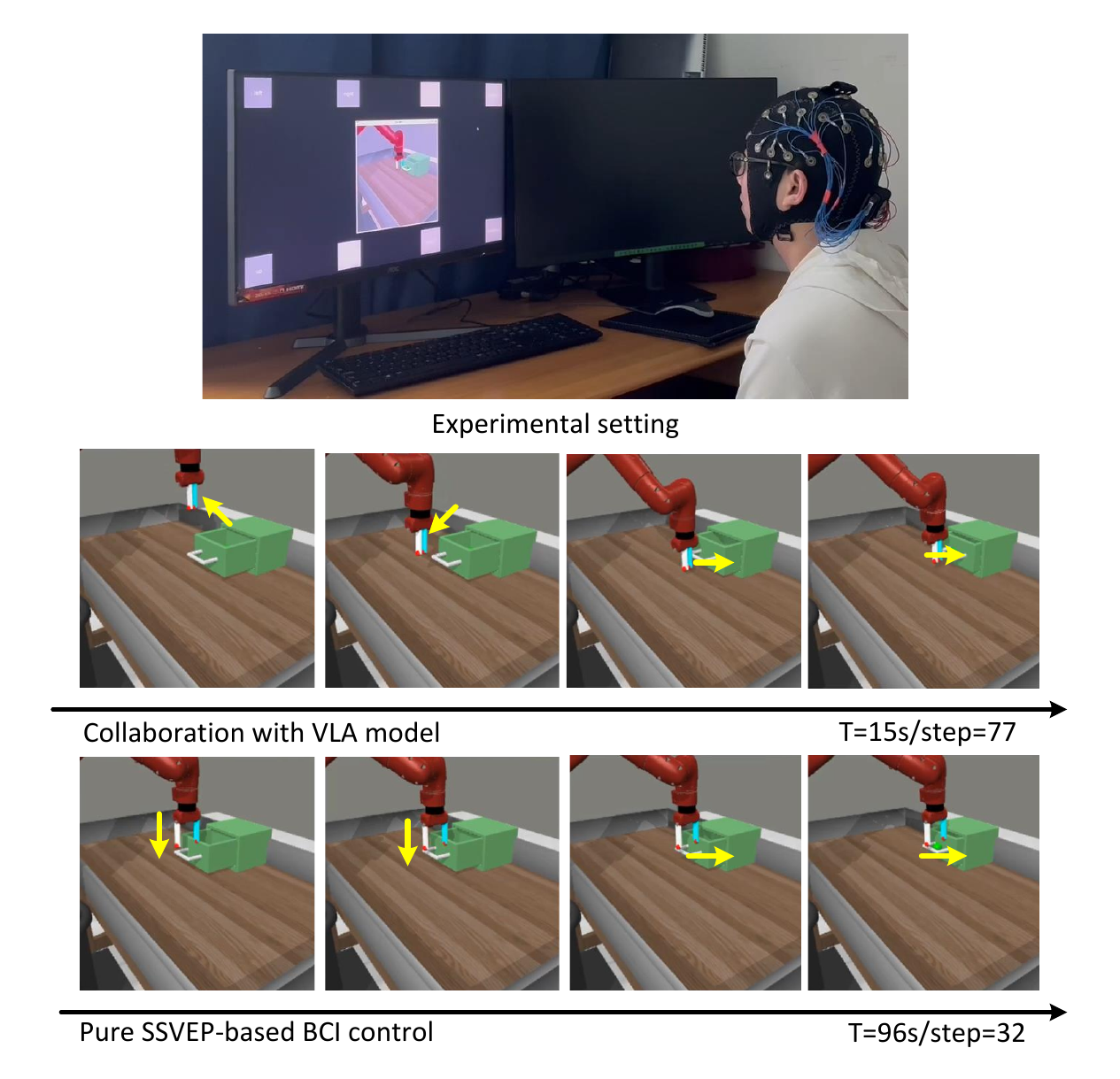}
       \caption{Application of the collaboration framework in SSVEP-based BCI: A comparison between pure SSVEP-based control and the collaboration between the VLA model and the BCI user. Although in some cases the policy of the human participant performs better than the VLA model (steps: 77 vs. 32), the collaboration system significantly improves time efficiency for a given task (time: 15s vs. 96s). }
    \centering
    \label{fig:BCI}
\end{figure}

\subsubsection{Human Users Become More Skillful During the Collaboration}

During collaboration, human users become more familiar with the VLA system and may adjust their policy to improve performance with the VLA model. This analysis focuses on the challenge tasks in MT10 where the success rate is lower than average (`pick place', `push', `peg insert', `reach'). As shown in Fig.~\ref{fig:human_learning}, this adaptation is observed across different users in the challenge tasks during the first five rounds of interaction with the VLA model. The average success rate shows a strong positive correlation with the number of interaction rounds (Pearson correlation: 0.95). Success rates improve from below average to above average. Similarly, the number of action steps taken by human users to complete tasks shows a negative linear correlation with rounds (Pearson correlation: -0.63). Initially, action steps are higher than average but decrease over time. Since the VLA model remains unchanged in these settings, this improvement is attributed to learning by the participants.

\begin{table}[!tb] \small
\renewcommand{\arraystretch}{1.2} 
\centering
\caption{Preliminary results of the application of the collaboration system VLA model in SSVEP-based BCI ($N=16$).}
\begin{tabular}{p{1.8cm} p{1.6cm}<{\centering} p{0.8cm}<{\centering} p{1.6cm}<{\centering} p{0.8cm}<{\centering}} 
\toprule

\multirow{2}{*}{\textbf{Tasks}} & \multicolumn{2}{c}{\textbf{Participant 1}} & \multicolumn{2}{c}{\textbf{Participant 2}} \\
\cmidrule{2-5}
& success rate & steps & success rate & steps \\
\midrule
window open  & 1.00 (5/5) & 100.20 & 1.00 (5/5) & 160.75 \\
drawer close & 1.00 (5/5) & 66.40 & 1.00 (5/5) & 79.80 \\
button press & 1.00 (5/5) & 82.60 & 1.00 (5/5) & 67.80 \\
door open & 1.00 (5/5) & 83.60 & 1.00 (5/5) & 83.80 \\
\bottomrule
\end{tabular}

\label{tbl:BCI}
\end{table}

\section{Discussion}

The proposed expert-VLA collaboration system has been empirically validated in the SSVEP-based BCI system. The SSVEP-based BCI paradigm typically requires a sustained period of visual stimulation to evoke a steady response in the brain, as reflected by EEG signals for decoding. In this context, the input action signal from the human participant (expert) is slow, limiting the system's responsiveness. However, by leveraging the VLA model for most actions, the proposed collaboration system enhances the whole system's ability to respond at higher speeds. This improvement not only increases efficiency but also significantly reduces the user's workload.

The EEG cap used in the experiment is the Emotiv Flex, collecting signals at 128 Hz, with electrodes near the occipital lobe selected for decoding. The decoding algorithm employed is Canonical Correlation Analysis (CCA). As this study is not focused on the BCI system itself, the experiment primarily aims to validate the feasibility of the VLA model-expert collaboration system. 

Two participants are involved in this preliminary experiment, performing four different tasks in the MT10 benchmark. The results are presented in Table~\ref{tbl:BCI}, showing that both participants can complete the tasks with the collaboration of the VLA model. Additionally, participants were asked to complete the manipulation task using pure SSVEP-based BCI control. Although the total number of steps may be reduced in some tasks, a significant reduction in time is achieved through collaboration with the VLA model, as it performs most actions at a much faster speed compared to the BCI system input (see the supplementary video and Fig.~\ref{fig:BCI}).

\section{Conclusion}

This study presents a collaboration framework between VLA models and experts for bi-directional manipulation learning. During manipulation, the proposed framework benefits both VLA models and experts by improving VLA performance while reducing the workload of human experts. Beyond manipulation, the framework facilitates bidirectional learning: the VLA model can be re-fine-tuned using manipulation data, while human experts improve their skills through interaction. This work provides a novel perspective on human-machine interaction, offering an effective approach to enhancing the efficiency of low-frequency human action input systems. Furthermore, it enables continuous improvement of VLA model performance through real-world application. Future work will focus on fine-tuning VLA models through online interaction and deploying the system in real robotic environments.

{\small
\bibliographystyle{IEEEtran}
\bibliography{mybib}
}

\end{document}